\documentclass[journal,twoside,web]{ieeecolor}
\pdfoutput=1
\usepackage{tmi}
\usepackage{cite}
\usepackage{amsmath,amssymb,amsfonts}
\usepackage{algorithmic}
\usepackage{graphicx}
\usepackage{textcomp}
\usepackage{algorithm}
\usepackage{algorithmic}
\usepackage{times}
\usepackage{courier}
\usepackage[hyphens]{url} 
\usepackage{makecell}
\usepackage{graphicx}
\floatstyle{ruled}
\newfloat{listing}{tb}{lst}{}
\floatname{listing}{Listing}
\usepackage{newfloat}
\usepackage{listings}
\usepackage{amsmath}
\usepackage{amssymb}
\usepackage{booktabs}
\usepackage{url}
\usepackage{multirow}

\newcommand{\cut}[1]{}

\newif\ifshowcomments
\showcommentstrue

\ifshowcomments

\newcommand{\TODO}[1]{{\color{red}{[TODO: #1]}}}

\else
\newcommand{\TODO}[1]{}
\newcommand{\revised}[1]{\color{blue}}
\fi

\makeatletter
\DeclareRobustCommand\onedot{\futurelet\@let@token\@onedot}
\def\@onedot{\ifx\@let@token.\else.\null\fi\xspace}

\makeatother

\def\BibTeX{{\rm B\kern-.05em{\sc i\kern-.025em b}\kern-.08em
    T\kern-.1667em\lower.7ex\hbox{E}\kern-.125emX}}
\begin{document}

\title{Exploring Segment-level Semantics for \\ Online Phase Recognition from Surgical Videos}
\author{Xinpeng Ding and Xiaomeng Li,  \IEEEmembership{Member, IEEE}

\thanks{Manuscript received XX XX, 2021.
This work was supported in part by the research grant from Shenzhen Municipal Central Government Guides Local Science and Technology Development Special Funded Projects under Grant 2021Szvup139 and in part by a research grant from HKUST-BICI Exploratory Fund (HCIC-004). (\emph{Corresponding author: Xiaomeng Li.})
}
\thanks{Xinpeng Ding is with the Department of Electronic and Computer Engineering, The Hong Kong University of Science and Technology,
Hong Kong, SAR, China (e-mail: xdingaf@connect.ust.hk).}
\thanks{Xiaomeng Li is with the Department of Electronic and Computer Engineering, The Hong Kong University of Science and Technology, Hong Kong, SAR, China, and also with The Hong Kong University of Science and Technology Shenzhen Research Institute, Shenzhen 518057, China (e-mail: eexmli@ust.hk).}
}
\maketitle


%
\begin{abstract}
Automatic surgical phase recognition plays a vital role in robot-assisted surgeries. Existing methods ignored a pivotal problem that surgical phases should be classified by learning \textit{segment-level semantics} instead of solely relying on frame-wise information. 
This paper presents a segment-attentive hierarchical consistency network (SAHC) for surgical phase recognition from videos. The key idea is to extract hierarchical high-level semantic-consistent segments and use them to refine the erroneous predictions caused by ambiguous frames. To achieve it, we design a temporal hierarchical network to generate hierarchical high-level segments. 
Then, we introduce a hierarchical segment-frame attention module to capture relations between the low-level frames and high-level segments. By regularizing the predictions of frames and their corresponding segments via a consistency loss, the network can generate semantic-consistent segments and then rectify the misclassified predictions caused by ambiguous low-level frames. We validate SAHC on two public surgical video datasets, \emph{i.e.}, the M2CAI16 challenge dataset and the Cholec80 dataset. \textcolor{black}{Experimental results show that our method outperforms previous state-of-the-arts and ablation studies prove the effectiveness of our proposed modules.}
%
\textcolor{black}{Our code has been released at: \url{https://github.com/xmed-lab/SAHC}.}

\begin{IEEEkeywords}
Surgical video analysis, surgical phase recognition.
\end{IEEEkeywords}

\end{abstract}
%
\section{Introduction}
\label{sec:introduction}
\begin{figure}[!t]
    \centering
    \includegraphics[width=0.9\columnwidth,height=0.3\textheight]{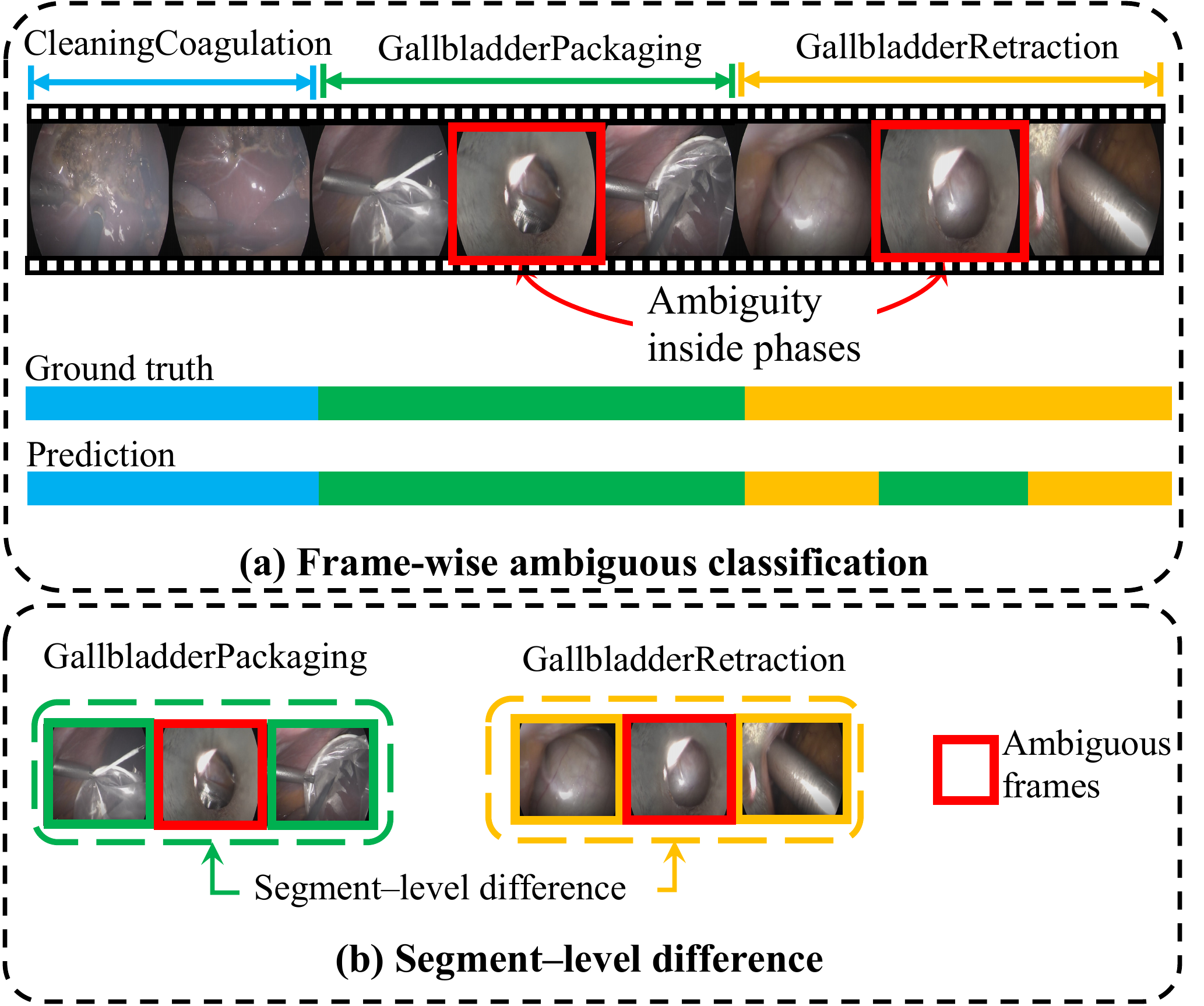}
    \caption{(a)
    %
    Ambiguous frame information would make the model generate erroneous predictions for surgical phase recognition.  
    %
    (b) Segment-level difference.
    Compared to the frame-level information, segment-level information can present more discriminative features for surgical phase recognition. 
    }
    \label{fig:motivation}
\end{figure}

Computer-assisted surgery systems can be used for pre-operative planning, surgical navigation and assist doctors in performing surgical procedures in modern operating rooms~\cite{maier2017surgical,tanwani2020motion2vec}. 
Surgical phase recognition, one crucial and challenging task in computer-assisted surgery systems, aims to recognize the surgical activities from a surgery video~\cite{twinanda2016endonet}.
It can monitor surgical processes and early alert the potential deviations and anomalies, advise the optimal arrangement, and provide clinical decision support~\cite{padoy2019machine}.
Hence, developing an accurate online phase recognition algorithm, \emph{i.e.}, predicts what phase is occurring at the current frame without knowing the future information from surgery videos, is highly demanded in clinical practice. 

%
%

%
%

One of the most critical challenges in automatically recognizing the surgical phase from videos is that the complicated surgical scenes usually have limited inter-phase variance and high intra-phase variance. 
As shown in Fig.~\ref{fig:motivation}(a), two frames from two different surgical phases, \emph{e.g.}, ``GallbladderPackaging'' and ``GallbladderRetraction'', has a high visual similarity, which may easily lead to misclassification of the surgical phases. 
Hence, learning the long-range temporal dynamics by observing the neighboring frames is the key solution to this problem.


Recent studies solved this issue by  capturing the long-term frame-wise relation between the current frame and previous frames~\cite{jin2021temporal,gao2021trans}.
For example, Jin~\emph{et al.}~\cite{jin2021temporal} introduced TMRNet, which contains a memory bank to store long-range information to learn the relation between the current frame and previous frames. 
%
%
Another \textcolor{black}{category of research}~\cite{czempiel2020tecno,yi2021not} employed a multi-stage architecture, including a predictor stage to generate a frame-wise prediction and a refinement stage to refine the previous prediction. 
For example, Czempiel~\emph{et al.}~\cite{czempiel2020tecno} introduced MS-TCN~\cite{farha2019ms} into surgical phase recognition, and used causal temporal convolutional networks~\cite{lea2017temporal} for online prediction. 
To solve the insufficient training of the refinement stage, Yi~\emph{et al.}~\cite{yi2021not} proposed a non-end-to-end stage to train the refinement stage separately.

Although these methods attempted to learn long-range temporal dynamics, they generated phase predictions by learning inter-frame relations in a low-level fine-grained way.
 \textcolor{black}{In this paper, we aim to explore that: would the frame-level phase recognition be improved by incorporating segment-level modeling?}
As shown in Fig.~\ref{fig:motivation}(b), two ambiguous frames (solid red boxes) are too similar, making the model hard to distinguish between two surgical phases. In contrast, their corresponding high-level segment (dashed boxes) can present discriminative semantics for surgical activities, contributing to the better recognition of surgical phase; \textcolor{black}{see Fig.~\ref{fig:seglevel} for details.}

To this end, we present a segment-attentive hierarchical consistency network (SAHC) for online phase recognition from surgical videos.
\textbf{\emph{The key idea is to learn high-level segments from surgical videos and then adopt them to rectify the ambiguous semantics caused by low-level frames.}}
To achieve it, we first design a hierarchical temporal feature extractor to generate high-level segments by capturing the feature of the current frame and its multi-scale neighboring frames.
Then, to rectify the ambiguous information in low-level frames, we develop a hierarchical segment-frame attention to capture the relations between the high-level temporal segments and low-level frames. 
%
By enhancing the consistency of the predictions from frames and their neighboring segments, we find that the features of ambiguous frames and their corresponding high-level segments would be pulled together, resulting in the effective refinement of the frame-wise ambiguity. 

This paper has the following contributions:
\begin{itemize}
     \item  \textcolor{black}{We introduce the importance of high-level segment information for surgical phase recognition, and leverage its semantics to refine the ambiguous low-level frames.}
     
    \item  We present a segment-attentive hierarchical consistency network (SAHC), which generates high-level semantic-consistent segments and then rectifies prediction errors via a segment-attentive module. 
    
    
     \item 
     We propose a hierarchical segment-frame attention module to learn relationships between frames and segments. 
     
    
    
    
    
    \item Experiments on two public surgical phase recognition datasets show that our method achieves a significant improvement over the prior art, \emph{e.g.}, over 3.8\% on the M2CAI16 dataset.
\end{itemize}

\begin{figure*}
    \centering
    \includegraphics[width=1.9\columnwidth,height=0.33\textheight]{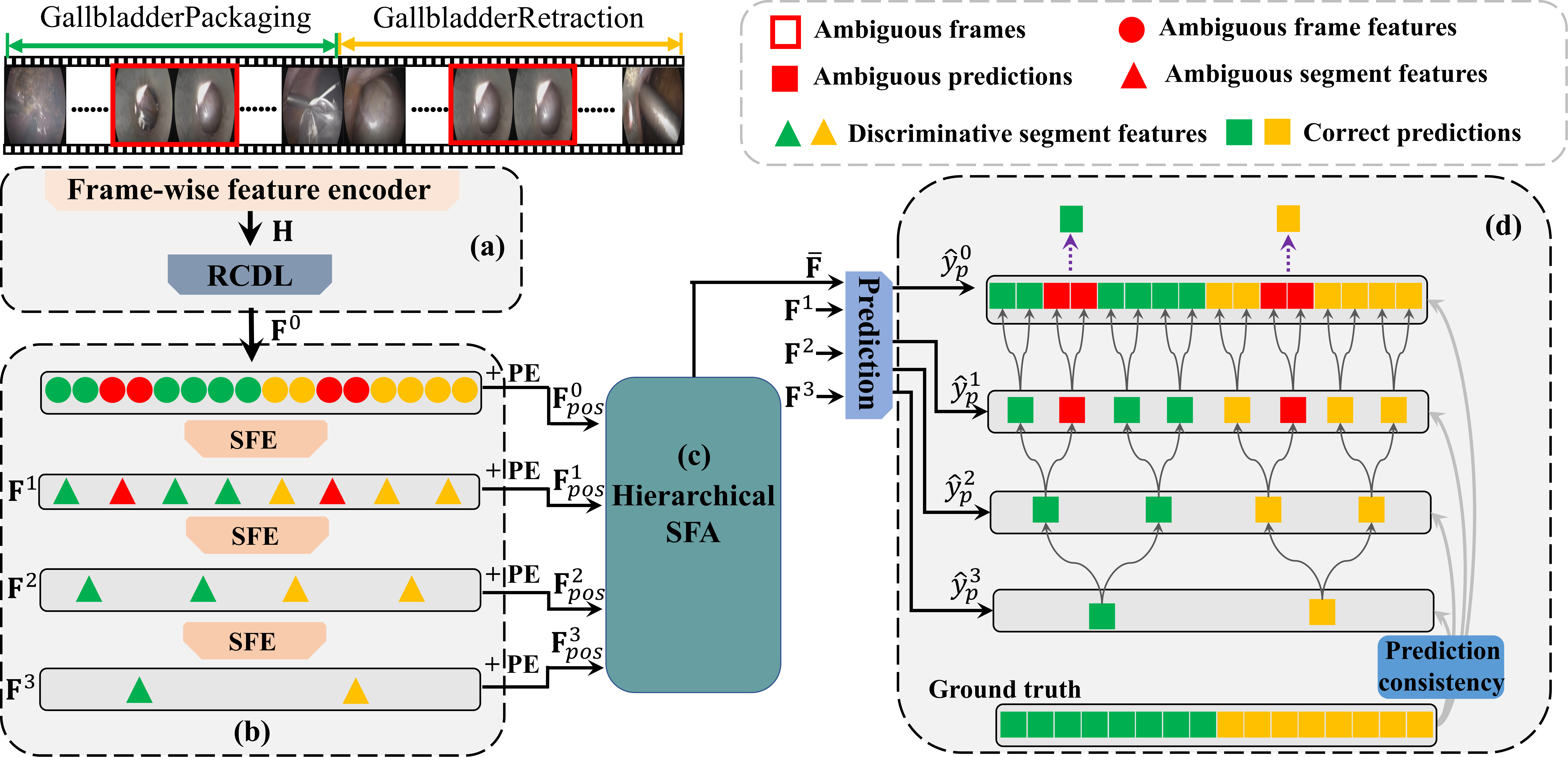}
    \caption{
    Illustration of our proposed Segment-Attentive Hierarchical Consistency Network (SAHC).
    The frame-wise feature $\mathbf{H}$ is extracted after feeding video frames into the frame-wise feature encoder. Then, several stacked residual causal dilated temporal convolution layers (RCDL)\textcolor{black}{, defined in Section~\ref{sec:spaital} and Fig.~\ref{fig:sfe}(b),} takes $\mathbf{H}$ as the input to obtain $\mathbf{F}^0$.
   Since the frames in different phases show very similar appearance,~\emph{i.e.}, ambiguous (see red hollow boxes), the model would extract some ambiguous features (\emph{i.e.}, red solid circles), which would result in ambiguous predictions (see red solid boxes).
    \textcolor{black}{To alleviate this, based on the frame features, we introduce $M$ segment-wise feature extractors (SFE)\textcolor{black}{, defined in Section~\ref{sec:sfe} and Fig.~\ref{fig:sfe},} to extract a set of hierarchical segment features $\{ \mathbf{F}^i \}_{i=1}^{M}$, which contains more discriminative information (see green and orange triangles).}
    After adding the positional encoding (PE) to $\{ \mathbf{F}^i \}_{i=1}^{M}$, we obtain $\{ \mathbf{F}^i_{pos} \}_{i=1}^M$ and then use a hierarchical segment-frame attention (SFA)\textcolor{black}{, defined in Section~\ref{sec:sfa} and Fig.~\ref{fig:mtt},} to capture the relations between frames and segments.
    \textcolor{black}{Finally, we regularize the network to encourage all prediction from low-level to high-level,~\emph{i.e.}, $\{ \hat{y}^i_p  \}_{i=0}^{3}$, to be consistent, which uses correct high-level predictions to refine the low-level ones.}
    }
    \label{fig:architecture}
\end{figure*}
%
\section{Related Work}
\subsection{Surgical Phase Recognition from Surgical Video}
Early work for surgical phase recognition from surgical videos is mainly based on hand-crafted features, such as pixel values and intensity gradients \cite{blum2010modeling}, spatial-temporal features \cite{zappella2013surgical} and features consisting of color, texture, and shape \cite{lalys2011framework}.
Concurrently, there are also other works that using linear statistical models to capture the temporal information of surgical videos,~\emph{e.g.}, left-right HMM~\cite{lalys2011framework}, Hidden
semi-Markov Model~\cite{dergachyova2016automatic}, hierarchical HMM~\cite{twinanda2016endonet}, conditional Random Fields~\cite{tao2013surgical,quellec2014real,lea2015improved} and Dynamic Time Warping~\cite{blum2010modeling}.
However, their performance is limited by the empirically designed low-level features.
In recent years, neural networks have extracted spatial and temporal features for surgical phase recognition from surgical videos. These methods can be broadly classified into two categories. 

%

One category aims at modeling spatial and temporal features with frame-wise labels. 
For example, Twinanda~\emph{et al.}~\cite{twinanda2016endonet} employed ResNet~\cite{he2016deep} to extract video-level features and demonstrated its effectiveness for surgical phase recognition.
Jin~\emph{et al.}~\cite{jin2017sv} introduced SV-RCNet, a unified framework that integrates ResNet and LSTM module to sequentially learn spatial-temporal features for surgical phase recognition.
%
%
To effectively capture the long-range temporal dynamics, Jin~\emph{et al.}~\cite{jin2021temporal} developed TMRNet, which consists of a memory bank to fuse long-range and multi-scale temporal features for surgical phase recognition. 
In addition to learning temporal information, Gao~\emph{et al.}~\cite{gao2021trans} used a hybrid embedding aggregation transformer to enhance the importance of spatial features for phase recognition.
Some work employed multi-stage architecture, \emph{e.g.}, a predictor stage and a refinement stage, in surgical phase recognition, where the misclassification in the predictor stage can be well rectified during the refinement stage.
For example, TeCNO~\cite{czempiel2020tecno} adapted the  multi-stage temporal convolution network (MS-TCN)~\cite{farha2019ms,li2020ms} to online surgery scenario via causal and dilated convolutions.
Yi~\emph{et al.}~\cite{yi2021not} found that directly using MS-TCN brings little improvement and then proposed a new non end-to-end training strategy.

Another category leverages additional information,~\emph{e.g.}, designing multi-task learning to improve the performance for surgical phase recognition. 
For instance, Twinanda~\emph{et al.}~\cite{twinanda2016endonet} trained a shared work for feature extraction in a multi-task way consisting of tool presence detection and phase recognition.
Zisimopoulos~\emph{et al.}~\cite{zisimopoulos2018deepphase} used a ResNet~\cite{he2016deep} to predict the binary predictions for the tool presence and then combine the predictions and features for phase recognition.
MTRCNet-CL \cite{jin2020multi} introduced a novel correlation loss to explicitly model the relations between tool presence and phase classification.
In addition to the tool information, Nakawala~\emph{et al.}~\cite{nakawala2019deep} leveraged more cues, such as management tools, ontology and production rules to improve the performance.
Some works~\cite{sarikaya2018joint} also extracted the optical flows and leverage the motion information to enhance the learning of the model.
These methods suffer from extra annotation cost for multi-tasking
or bring additional computation overhead to obtain other modalities,~\emph{e.g.}, optical flows.

Our method belongs to the first category. Most existing methods aim to capture frame-wise relations by learning temporal dynamics while ignoring the high-level semantic information and its influences for low-level frames. 
\textcolor{black}{There are some works~\cite{quellec2014real} using conditional Random Fields to combine frame-level information for high-level surgical tasks.
However, in surgical phase recognition, we need detailed frame-wise classification.
}
\textcolor{black}{In this paper, we combine the high segment-level and low frame-level semantics for surgical phase recognition.
By the proposed segment-frame attention, our model uses high-level semantics to refine low-level errors to improve the frame-level performance.
}

\if 1 
However, most of these methods aim to capture frame-wise relations, while ignoring the semantic consistency of high-level temporal segments in intra-phases.
%
In this paper, for the first time, we use the high-level segment information to refine the fine-grained frame-wise one.
\fi 

\subsection{Temporal Pyramid Learning in Videos}


Our method is also related to the temporal pyramid learning in video action recognition tasks.
Temporal pyramid methods aim to process the variant duration of actions by extracting multi-scale temporal information from videos.

Early approaches generated a fixed multi-scale sliding windows, which act as proposals for temporal action localization \cite{shou2016temporal,ding2020weakly,li2021multi} or video grounding \cite{gao2017tall,anne2017localizing,Ding_2021_ICCV,zeng2020dense}.
Recently, researchers sampled frames at different temporal rates to construct an input-level frame pyramid.
And frames in each level of the pyramid were extracted by separate networks to obtain the corresponding mid-level features, which were then fused for final prediction~\cite{zhou2018temporal,feichtenhofer2019slowfast}.
%
%
However, these methods required the additional networks, which may be computationally expensive~\cite{yang2020temporal}.
Motivated by \textcolor{black}{the feature pyramid network (FPN)}~\cite{lin2017feature}, Yang~\emph{et al.}~\cite{yang2020temporal} captured visual tempos in multi-scale feature levels with only a single input. 
Specifically, they utilized a feature pyramid network that temporally downsamples features to obtain different temporal scales.


Compared with existing temporal pyramid networks~\cite{yang2020temporal,zeng2020dense} for video analysis in computer vision, our method has the following differences.
(1) Different video tasks and bottlenecks. Existing methods~\cite{yang2020temporal,zeng2020dense} proposed to learn multi-temporal scales to tackle the variant duration of the action instances for natural video action recognition. In contrast, our goal is to use segment-level information to \textbf{\emph{refine the erroneous predictions caused by ambiguous frame-level information in surgical videos}}.  
(2) Network design with consistency regularization.
We design a hierarchical network with consistency regularization to rectify the erroneous prediction caused by ambiguous frame-level information.  
(3) Attention. We further introduce an attention module to capture the relationship between frame- and segment-level information. These two contributions do not exist in current temporal pyramid methods.  

\subsection{Long-term Video Understanding}
\textcolor{black}{
To capture long-term information of videos, Farha~\emph{et al.}~\cite{farha2019ms,li2020ms} propose a multi-stage temporal convolution network (MS-TCN) for the temporal action segmentation task, which enlarges the receptive fields to capture long-range temporal information by cascaded dilated 1D convolutions.
To reduce the annotation cost in long videos, Zhukov~\emph{et al.}~\cite{zhukov2020learning} introduces a long-range temporal order verification to  to isolate actions from their background in a self-supervised manner.
Wu~\emph{et al.}~\cite{wu2019long} propose a long-term feature bank to contain supportive information from the entire video, which augments the state-of-the-art video models that otherwise would only view short clips of 2-5 seconds.
Recently, Object Transformer~\cite{wu2021towards} is proposed to use transformer~\cite{vaswani2017attention} to model the long-term relations.
In this paper, we follow previous works~\cite{czempiel2020tecno,yi2019hard,yi2021not} to use cascaded TCN to capture long-range information for surgical videos.
}



\section{Methodology}
\if 1 
As the above analysis, too limited inter-variance in different phases would cause ambiguous classification.
%
%
Motivated by that the semantics in the same surgical phase are consistent and contiguous, hence, for the first time, we introduce high-level segments-level prediction and use it to refine the frame-level ambiguity by enhancing the local semantic consistency between the frame and its neighbour segments.

%
\fi 

Fig.~\ref{fig:architecture} illustrates our proposed segment-attentive hierarchical consistency network (SAHC).
%
The video frames are firstly fed into {\bf (a)} a spatial-temporal feature extractor to obtain the features with temporal frame-level information, followed by {\bf (b)} a segment-level hierarchical network consists of three segment-wise feature extractors (SFE) to capture the multi-scale segment-level semantics.
After that, we introduce {\bf (c)} a segment-frame attention (SFA) module to jointly learn the relations between frames and segments.
Finally, {\bf (d)} the segment-frame hierarchical consistency loss is developed to enhance the consistent prediction from low-level frames and their neighbouring segments, such that the erroneous frame-wise prediction can be rectified by segments. 
In the following sections, we describe our method in detail. 

\subsection{Spatial-temporal Feature Extractor}
\label{sec:spaital}

%
We denote a video as $\mathbf{V} = \{ \mathbf{v}_t \}_{t=0}^{T-1}$, where $T$ is the number of frames and $\mathbf{v}_t\in \mathbb{R}^{H \times W \times 3}$ is a frame with height $H$, weight $W$ and three channels.
Let $\mathcal{C}$ denote a set of surgical phases, where $\mathcal{C} = \{0,...,C-1\}$ and $C$ is the number of phase categories.
Our goal is to learn a \textcolor{black}{deep} network $f_{\theta}$ that maps the input  $\mathbf{v}_t$ to a phase label $y_t \in \mathbb{R}^{C}$, which is a one-hot vector of phase label $c \in \mathcal{C}$.

\begin{figure}
    \centering
    \includegraphics[width=1.0\columnwidth]{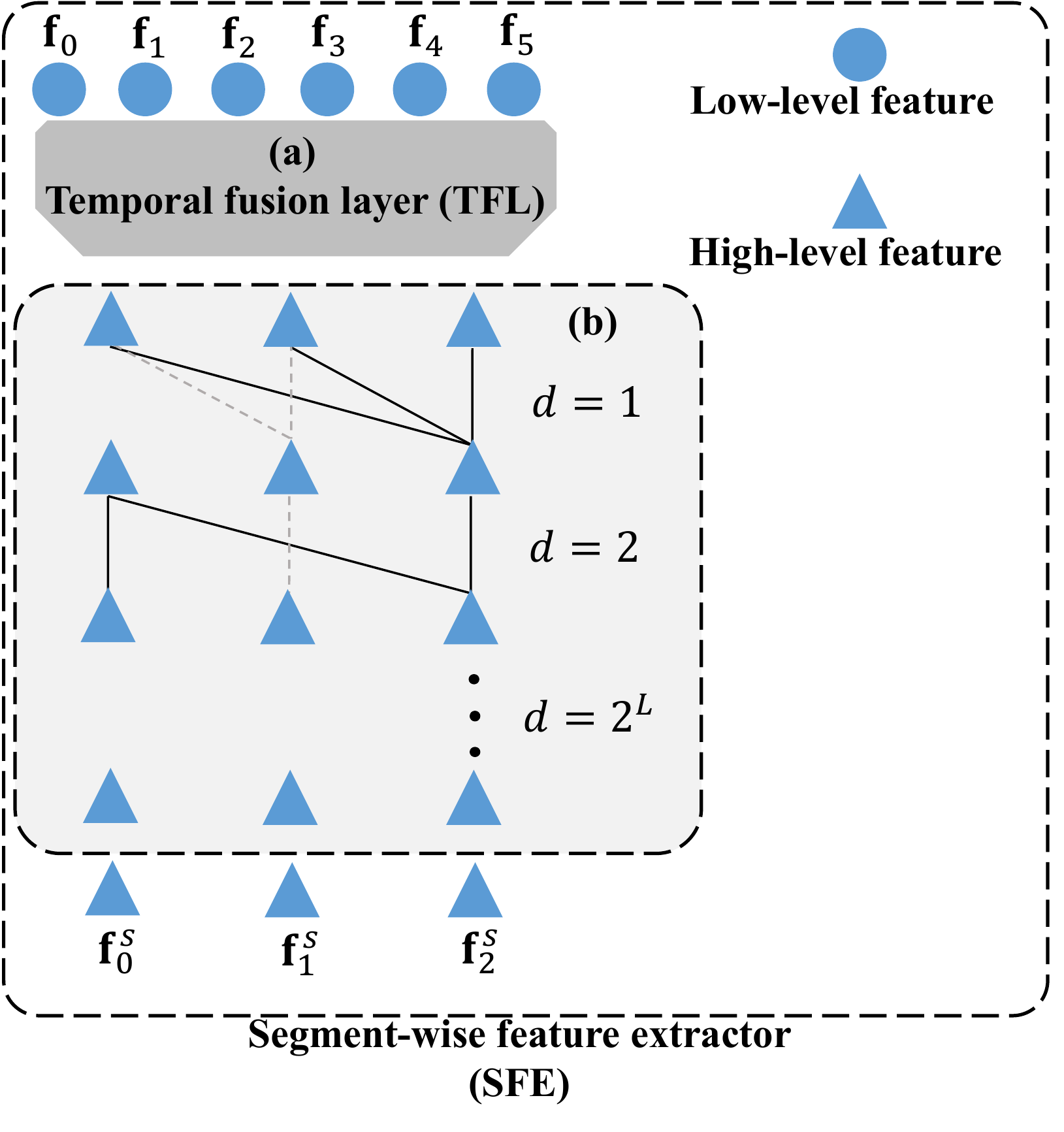}
    \caption{
    Illustration of SFE.
    It consists of two main components: (a) the temporal fusion layer and (b)  $L$ stacked residual causal dilated temporal convolution layers (RCDL). \textcolor{black}{After feeding low-level features into the temporal fusion layer, we obtain the high-level features. Then, we use RCDL to capture the relations between high-level features.}
    }
    \label{fig:sfe}
\end{figure}

In order to capture the spatial-temporal information of the videos, we first extract the frame-wise feature, and then use the temporal convolutional network to model their temporal relations.
Specifically, we first feed the frames $\mathbf{V}$ into the the spatial encoder, \emph{i.e.}, ResNet-50 \cite{he2016deep}, to extract the spatial feature of each frame, denoted as ${\mathbf{H}}=\{ {\mathbf{h}}_t\}^{T-1}_{t=0}$, where ${\mathbf{h}}_t \in \mathbb{R}^{ D}$ is the feature of the $\mathbf{v}_t$ and $D$ the dimension of the feature.
Then, we feed $\mathbf{H} \in \mathbb{R}^{T \times D} $ into several stacked residual causal dilated temporal convolution layers (RCDL) (shown in Fig.~\ref{fig:sfe}(b)) to capture the frame-wise relation and obtain the corresponding features $\mathbf{F}^0 \in \mathbb{R}^{T \times D}$.
The operations of each RCDL can be formally as follows:
\begin{equation}
    \mathbf{Z}_{l}= ReLU\left(\mathbf{W}_{1, l} * \mathbf{F}_{l-1}+\mathbf{b}_{1, l}\right),
\end{equation}
\begin{equation}
    \mathbf{F}_{l} = \mathbf{F}_{l-1}+\mathbf{W}_{2, l} * \mathbf{Z}_{l} + \mathbf{b}_{2, l},
    \label{E:residual}
\end{equation}
where $\mathbf{F}_{l}$ is the output of the layer $l$, $*$ denotes the convolution operator, $\mathbf{W}_{1, l}$ is the dilated 1D convolution kernel \cite{czempiel2020tecno}, $\mathbf{W}_{2, l}$ is the weights of a $1 \times 1$ convolution and $\mathbf{b}_{1, l}$, $\mathbf{b}_{2, l}$ are bias vectors.
The dimension of $\mathbf{F}^i$ is set to $64$ in all RCDL.
Following \cite{czempiel2020tecno}, to predict the label of frame $\mathbf{f}_t$, casual dilated convolution only relies on the current and previous frames, \emph{i.e.}, $\left(\mathbf{f}_{t-n}, \ldots, \mathbf{f}_{t}\right)$, which allows for the online recognition.
Before the residual addition in Eq.~\ref{E:residual}, we adopt the dropout \cite{srivastava2014dropout} to avoid over-fitting.
In this paper, we set $L$ to be $11$ as same in \cite{li2020ms}.
%
To achieve online recognition of surgical activities, we use the casual dilated convolution that only relies on the current and previous frames, \emph{i.e.}, $\left(\mathbf{f}_{t-n}, \ldots, \mathbf{f}_{t}\right)$. 

\subsection{Segment-level Hierarchical Network}
\subsubsection{Segment-level Feature Extractor \textbf{(SFE)}.}~\label{sec:sfe}
After obtaining the frame-level temporal features $\mathbf{F}^0$, we then use them to extract the information of segments by a segment-level feature extractor.
%
%
%
Each segment-level feature extractor consists of a temporal fusion layer (see Fig. \ref{fig:sfe}(a)), followed by several $L$ stacked RCDL (see details in Section~\ref{sec:spaital} and Fig.~\ref{fig:sfe}(b)).
The goal of the temporal fusion layer is to aggregate the features of a frame and its neighbouring frames to obtain their corresponding segment features, which can be formally defined as:
\begin{equation}
    \mathbf{f}^s_p = {\rm TFL}(\mathbf{f}^0_t, \ldots, \mathbf{f}^0_{t+k}),
    \label{e:tfl}
\end{equation}
where TFL refers to the operation of the temporal fusion layer,  $\mathbf{f}^s_p$ is the segment feature and $k$ indicates the number of frames that each segment contains.
We have several choices to devise the temporal fusion layer, \emph{e.g.}, a convolution layer, a max-pooling layer or an average pooling layer with kernel size and stride of $k$. 
For examples shown in Fig.~\ref{fig:sfe}, \textcolor{black}{given a sequence low-level input features $\{ \mathbf{f}_i \}_{i=0}^{5}$} and a temporal fusion layer with kernel size and stride of $2$, we will then obtain the \textcolor{black}{high-level output features of $\{ \mathbf{f}^s_i \}_{i=0}^{2}$}.
The segment feature $\mathbf{f}_{0}^s$ aggregates $\mathbf{f}_0$ and $\mathbf{f}_1$, and  $\mathbf{f}_{1}^s$ aggregates $\mathbf{f}_2$ and $\mathbf{f}_3$, and so on. 

Hence, after feeding $\mathbf{F}^0$ into the temporal fusion layer, we obtain the segment-level features, defined as $\mathbf{F}^s = \{ \mathbf{f}^s_p \}_{p=0}^{T^s-1} \in \mathbb{R}^{T^s \times D}$.
$T^s =\left \lfloor T/k \right \rfloor$ and $\left \lfloor \cdot \right \rfloor$ indicates the round down.
We can achieve this by several methods, \emph{e.g.}, the convolution, the max-pooling and the average-pooling with setting both the kernel size and stride to $k$. We will discuss these different kinds of temporal fusion layers in Section~\ref{sec:ablation_param}.
Similar in frame-level, we also expect the model to capture the temporal relations between segments. 
To this end, we input $\mathbf{F}^s$ into RCDL to use temporal convolution to capture relations in $\mathbf{F}^s$ and finally obtain the output ${\mathbf{F}^1}$.

The implementation of our temporal fusion layer,~\emph{i.e.}, fusing frames to obtain the segments, may generate the ambiguous segments, which contains frames belonging to different phases.
In surgical videos, the category of frames in the same phases are consistent. Hence, only segments in the boundaries between two phases would contain frames with different class labels.
That is to say, the number of ambiguous segments is too few and can be ignored. 
\begin{figure}
    \centering
    \includegraphics[width=1.0\columnwidth,height=0.22\textheight]{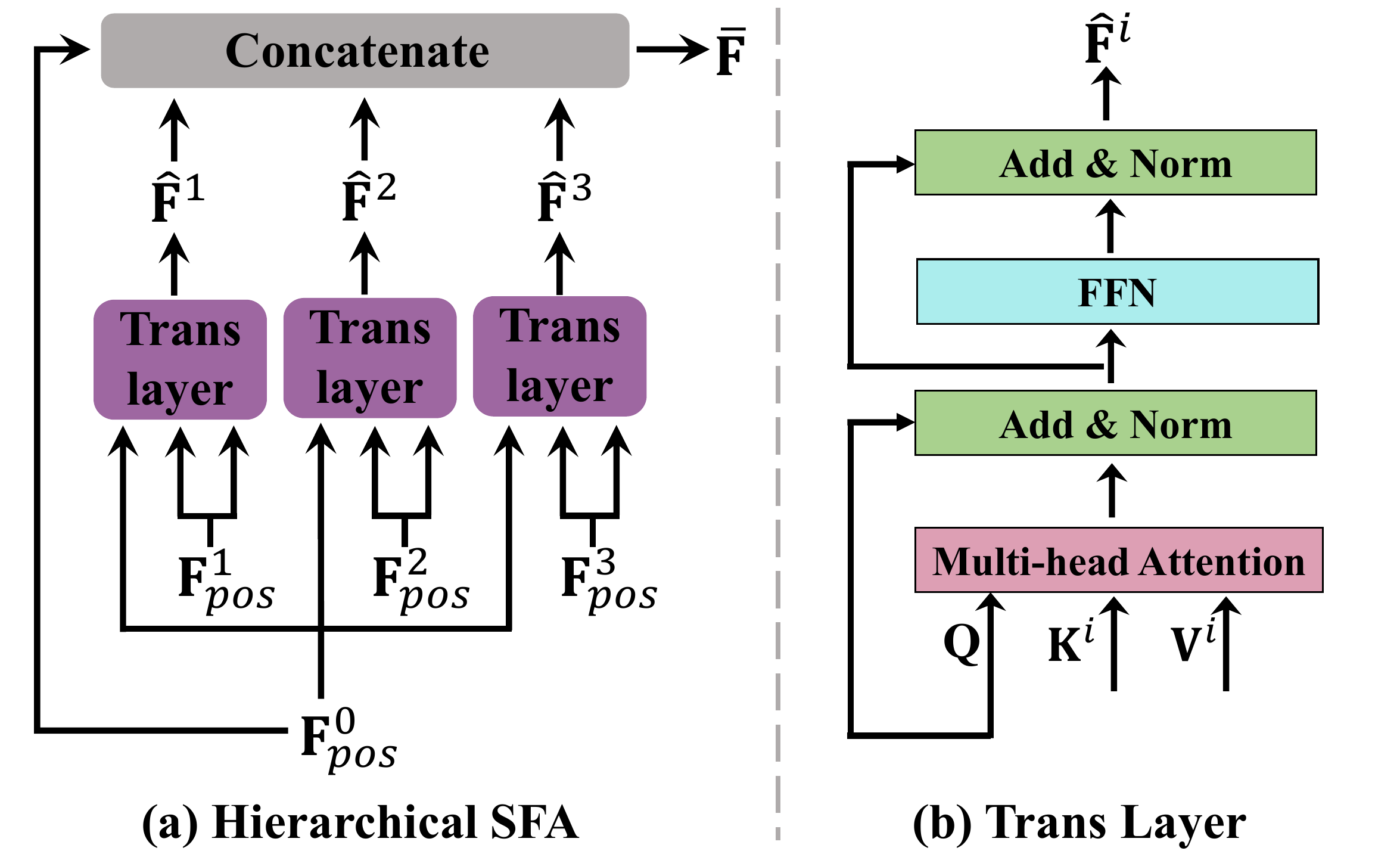}
    \caption{
    Illustration of (a) the hierarchical SFA (described in Fig.~\ref{fig:architecture}(c)) and (b) the transformer (Trans) layer.
    %
    %
    }
    \label{fig:mtt}
\end{figure}

\subsubsection{Hierarchical Network.}~\label{sec:hierarchical}
The duration of surgical videos and phases is various~\cite{jin2021temporal}; hence we need to extract the multi-scale temporal information for segments.
To achieve it, we introduce a hierarchical network, consisting of a sequence of segment-level feature extractor modules, to obtain a set of features,~\emph{i.e.}, $\{ \mathbf{F}^i \in \mathbb{R}^{T_i \times D} \}_{i=1}^{M}$, \textcolor{black}{where $M$ is the number of different  temporal scales}.
In this paper, we set $M=3$ empirically.
$\mathbf{F}^i$ is the output of the $i$-th segment-level feature extractor in the segment-level feature extractor sequence, and $T^i$ is its temporal duration and $T^{i+1} = \left \lfloor T^i /k \right \rfloor$.

\subsection{Segment-Frame Attention \textbf{(SFA)} Module}
\label{sec:sfa}
%
\textcolor{black}{The proposed RCDL in Section~\ref{sec:sfe} can only model the relations in each scale, \emph{i.e.}, the relations in $\mathbf{F}^i$.}
To use the high-level segment information to refine the erroneous predictions in low-level frames, \textcolor{black}{we need to capture the relations between features in different scales},~\emph{e.g.}, frames $\mathbf{F}^0$ and its corresponding segments $\mathbf{F}^1$ and $\mathbf{F}^2$ and $\mathbf{F}^3$, respectively.
%
%
%
Recently, Transformer~\cite{vaswani2017attention}, a kind of attention layer, shows promising results in learning attentive weights/relationships with applications in images, words, and videos \cite{vaswani2017attention,shao2020intra,dosovitskiy2020image,arnab2021vivit,liang2021swinir,liu2021swin,chen2021pre,sun2021rethinking}.
Here, we adopt the transformer to learn the relationships between segments and frames. 

%

To this end, we design a segment-frame attention (SFA) module, where frames are used as the queries, and hierarchical segments are served as keys and values.
In this way, the query,~\emph{i.e.}, a frame, can find its relation with segments overall the video across different phases, and use the features of its high-level segments to refine the ambiguous frames,\textcolor{black}{~\emph{i.e.}, pushing the frame-level features to be close with the segment-level ones. See qualitative analysis in Fig.~\ref{fig:att&sim}.}
Since there is no frame sequence information in the attention module, we need to embed position encoding additionally, which is formulated as: 
\begin{equation}
    \mathbf{F}^i_{pos}=\mathbf{F}^i + \mathbf{E}, i \in \{ 0, ..., M\}
\end{equation}
where $\mathbf{E} \in \mathbb{R}^{T \times D}$ is the learned positional embedding, \textcolor{black}{$M$ is the number of different temporal scales (defined in Section~\ref{sec:hierarchical})} and we set $M=3$ empirically.
Fig.~\ref{fig:mtt}(a) shows the details of the hierarchical SFA. 
%
Concretely, $\mathbf{F}^0_{pos}$ and $\{ \mathbf{F}^i_{pos} \}_{i=1}^3$ are fed into three shared transformer layers (Trans layer) and then generate a set of outputs, defined as $\{ \hat{\mathbf{F}}^i \}_{i=1}^3$.
We finally concatenate the multi-scale aggregated features $\{ \hat{\mathbf{F}}^i \}_{i=1}^{3}$ to obtain the final feature $\overline{\mathbf{F}} = \{ \overline{\mathbf{f}}_t \}_{t=0}^{T-1}$.
The details of the Trans layer are shown in Fig.~\ref{fig:mtt}(b).
The core component of the Trans layer is the multi-head attention, and the intuitive idea is that each token can interact with other tokens and can learn to aggregate useful semantics more effectively.
%
%
Given $\mathbf{F}^0_{pos}$ and $\mathbf{F}^i_{pos}$, each head of the multi-head attention can be formally defined as follows:
\begin{equation}
    \begin{aligned}
   \operatorname{Attention}\left(\mathbf{Q} ; \mathbf{K}^i ;\mathbf{V}^i\right) &=\operatorname{softmax}\left(\frac{ \mathbf{Q} {\mathbf{K}^i}^{\top}}{\sqrt{D}}\right) {\mathbf{V}^i} 
\end{aligned}
\end{equation}
where $\mathbf{Q}=\mathbf{F}^0_{pos}\mathbf{W}_{i}^{q}$, $\mathbf{K}^i=\mathbf{F}^i_{pos}\mathbf{W}_{i}^{k}$ and $\mathbf{V}^i=\mathbf{F}^i_{pos}\mathbf{W}_{i}^{v}$ are linear layers, and $\mathbf{W}_{i}^{q}, \mathbf{W}_{i}^{k}, \mathbf{W}_{i}^{v} \in \mathbb{R}^{D \times \frac{D}{N_{h e a d}}}$.
$N_{head}$ is the number of heads and $\sqrt{D}$ controls the effect of growing magnitude of dot-product with larger $D$ \cite{vaswani2017attention}.
Subsequently, the output of all heads are concatenated and fed into a linear layer, followed by a layer normalization \cite{ba2016layer}.
Then, it is followed by a feed-forward network with ReLU activation.
The residual connection \cite{he2016deep} and layer normalization \cite{ba2016layer} are also applied as in the multi-head attention.
Finally, we obtain the output of each Trans layer, $\hat{\mathbf{F}}^i$.
%
%
%
%
%
We then concatenate the multi-scale aggregated features $\{ \hat{\mathbf{F}}^i \}_{i=1}^{3}$ to obtain the final feature $\overline{\mathbf{F}} = \{ \overline{\mathbf{f}}_t \}_{t=0}^{T-1}$.
%
\subsection{Consistency between Frames and Segments}
Based on the final obtained features of frames $\mathbf{F}^0$ and multi-scale segments $\{ \mathbf{F}^i \}_{i=1}^3$,
we get their corresponding predictions $\hat{y}^0_t$ and $\hat{\mathcal{Y}}_t = \{ \hat{y}^i_p  \}_{i=1}^{3}$ by a shared prediction network,
where $\{ \hat{y}^0_t \}  \in \mathbb{R}^{C \times T}$ and $ \{ \hat{y}^i_p \in \mathbb{R}^{C \times T^i} \}_{i=1}^3$.
The shared prediction network is a convolution layer. The kernel size, stride, input dimension and the output dimension are 1, 1, 256, 7, respectively.
The network simply follows previous state-of-the-art methods~(Yi and Jiang 2021; Czempiel et al. 2020).
Then, the classification losses of the frames ($\mathcal{L}_{frame}$) and their corresponding neighbouring segments ($\mathcal{L}_{segment}$) can be defined as follows:
\begin{equation}
  \mathcal{L}_{frame} =  -\frac{1}{T} \sum_{t=0}^{T-1} \sum_{c =0}^{C-1} y^0_{t,c} \log (\hat{y}^0_{t,c}),
  \label{E:frame}
\end{equation}
\begin{equation}
  \mathcal{L}_{segment} =  - \frac{\beta}{T^i}\sum_{i=1}^{3}   \sum_{p=0}^{T^i-1} \sum_{c =0}^{C-1} y^i_{p,c} \log (\hat{y}^i_{p,c}),
  \label{E:seg}
\end{equation}
where $y_{p,c}^i \in \mathbb{R}^{C \times T^i}$  is the corresponding ground truth at $i$-th scale, which is generated by applying down-sampling to $y_{t,c}$.
$\beta$ is the hyper-parameter to control the weight of the segment-wise prediction at different scales.
The multi-scale consistency loss can be the combination of $\mathcal{L}_{frame}$ and $\mathcal{L}_{segment}$:
\begin{equation}
  \mathcal{L}_{msc} = \mathcal{L}_{frame} + \mathcal{L}_{segment}.
\end{equation}
In this way, the model would encourage the consistency of the prediction of the frames and their segments at multi-scale tempos.
For predicting more smoothing results, a mean squared error over the classification probabilities of every two adjacent frames are used \cite{farha2019ms}:
\begin{equation}
\begin{aligned}
   \mathcal{L}_{smooth} = & \frac{1}{T C} \sum_{c=0}^{C-1} \sum_{t=1}^{T-1}\left|\hat{y}^0_{t,c}-\hat{y}^0_{t+1, c}\right|^{2} \\ &+  \frac{\beta}{T^i C} \sum_{i=1}^{3}   \sum_{p=0}^{T^i-1} \sum_{c =0}^{C-1} \left|\hat{y}^i_{p,c}-\hat{y}^i_{p+1, c}\right|^{2}.
   \end{aligned}
\end{equation}
Hence, the overall of the objective of our SAHC is:
\begin{equation}
    \mathcal{L} = \mathcal{L}_{msc} + \lambda \mathcal{L}_{smooth},
    \label{E:all}
\end{equation}
where $\lambda$ is a hyper-parameter to control the weight between two losses. %
We will discuss the effect of $\alpha$ and $\lambda$ in Experiments.

\section{Experimental Results}
In this section, we evaluate our proposed SAHC on two benchmark datasets for surgical phase recognition. 
\begin{table}[t]
\centering
\caption{Comparisons with the state-of-the-arts on M2CAI16 Dataset.}\smallskip
\label{table:m2cai}
\resizebox{1.0\columnwidth}{!}{
\begin{tabular}{c| c c c c }
   Methods & Accuracy & Precision & Recall & Jaccard\\
\Xhline{1.5pt}
  PhaseNet~\cite{twinanda2016single}  &  \textcolor{black}{79.5 $\pm$ 12.1} &\textcolor{black}{-} & \textcolor{black}{-} & \textcolor{black}{64.1 $\pm$ 10.3}\\
  SV-RCNet~\cite{jin2017sv} & \textcolor{black}{81.7 $\pm$ 8.1} & \textcolor{black}{81.0 $\pm$ 8.3} &\textcolor{black}{81.6 $\pm$ 7.2} & \textcolor{black}{65.4 $\pm$ 8.9}\\
  OHFM~\cite{yi2019hard} & \textcolor{black}{85.2 $\pm$ 7.5} & \textcolor{black}{-} & \textcolor{black}{-} & \textcolor{black}{68.8 $\pm$ 10.5} \\
  TMRNet~\cite{jin2021temporal} &\textcolor{black}{87.0 $\pm$ 8.6} & \textcolor{black}{87.8 $\pm$ 6.9} & \textcolor{black}{88.4 $\pm$ 5.3} & \textcolor{black}{75.1 $\pm$ 6.9} \\
 Not-End~\cite{yi2021not} & \textcolor{black}{84.1 $\pm$ 9.6} & \textcolor{black}{-} &  \textcolor{black}{88.3 $\pm$ 9.6} & \textcolor{black}{69.8 $\pm$ 10.7} \\
  Trans-SVNet~\cite{gao2021trans} & 87.2 $\pm$ 9.3 &  88.0 $\pm$ 6.7 & 87.5 $\pm$ 5.5 & 74.7 $\pm$ 7.7 \\
 \hline 
 Ours & \textcolor{black}{{\bf 91.6 $\pm$ 7.8}} & \textcolor{black}{{\bf 93.5 $\pm$ 5.6}} & \textcolor{black}{{\bf 92.9 $\pm$  6.3}} & \textcolor{black}{{\bf 85.7 $\pm$  7.7}}\\
\end{tabular}}
\end{table} 
\begin{table}
\centering
\caption{Comparisons with the state-of-the-arts on Cholec80 Dataset.}\smallskip
\label{table:chole80}
\resizebox{1.0\columnwidth}{!}{
\begin{tabular}{c| c c c c }
   Methods & Accuracy & Precision & Recall & Jaccard\\
\Xhline{1.5pt}
   PhaseNet~\cite{twinanda2016single}  & 78.8 $\pm$ 4.7 & 71.3 $\pm$ 15.6 & 76.6 $\pm$ 16.6 & - \\
   SV-RCNet~\cite{jin2017sv} & 85.3 $\pm$ 7.3 & 80.7 $\pm$ 7.0 & 83.5 $\pm$ 7.5 & -\\
   OHFM~\cite{yi2019hard} & 87.3 $\pm$ 5.7 & - & - & 67.0 $\pm$ 13.4 \\
   TeCNO~\cite{czempiel2020tecno} & 88.6  $\pm$ 2.7 & - & 85.2  $\pm$ 10.6  & - \\
   TMRNet~\cite{jin2021temporal} & 90.1 $\pm$ 7.6 & 90.3 $\pm$ 3.3 & 89.5 $\pm$ 5.0 & 79.1 $\pm$ 5.7 \\
 Not-End~\cite{yi2021not} & \textcolor{black}{88.8 $\pm$ 7.1} & \textcolor{black}{-} &  \textcolor{black}{84.9 $\pm$ 7.2} & \textcolor{black}{73.2 $\pm$ 9.8}\\
 Trans-SVNet~\cite{gao2021trans} & 90.3 $\pm$ 7.1 &  90.7 $\pm$ 5.0 & 88.8 $\pm$ 7.4 & 79.3 $\pm$ 6.6 \\
 \hline 
 Ours & \textcolor{black}{{\bf 91.8 $\pm$ 8.1}} &  \textcolor{black}{{\bf 90.3 $\pm$ 6.4}} & \textcolor{black}{{\bf 90.0 $\pm$  6.4}} & \textcolor{black}{{\bf 81.2 $\pm$  5.5}}\\
\end{tabular}}
\end{table} 
\begin{figure}[t]
    \centering
    \includegraphics[width=1.0\columnwidth]{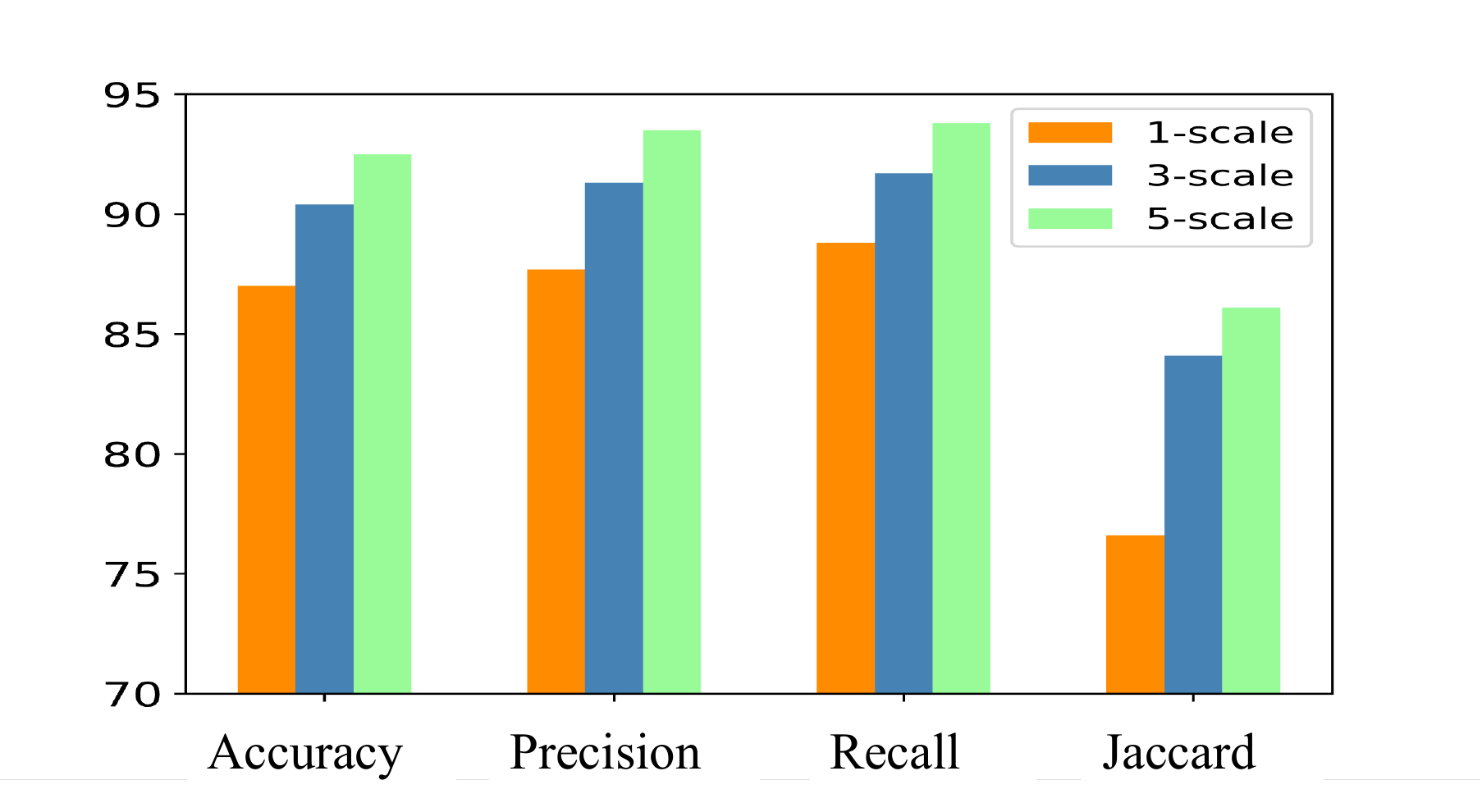}
    \caption{
    \textcolor{black}{Effect of learning with segment-level semantics. We visualize results, \emph{i.e.}, Accuracy, precision, recall and jaccard (see definition in Section~\ref{sec:datasets}), on baseline model with different temporal scale, \emph{i.e.}, $1,3,5$, on the M2CAI16 dataset~\cite{stauder2016tum}. It is clear that segments with more frames achieve higher performance.}
    }
    \label{fig:seglevel}
\end{figure}
\subsection{Datasets, Metrics and Implementation Details}~\label{sec:datasets}

\noindent\textcolor{black}{{\bf Datasets.}}
{\bf (a) M2CAI16.}
The M2CAI16 dataset~\cite{stauder2016tum} consists of $41$ laparoscopic videos at $25$ fps and each frame is with a resolution of $1920 \times 1080$.
The videos contain $8$ surgical phases, \emph{i.e.}, \emph{TrocarPlacement, Preparation, CalotTriangleDissection, ClippingCutting, GallbladderDissection, GallbladderRetraction, CleaningCoagulation, GallbladderPackaging,} labeled by professional surgeons.
More detailed information can be found in \cite{jin2017sv}.
\textcolor{black}{We use $27$ videos for training, $7$ videos for validation and $14$ videos for testing, following the same evaluation protocols in~\cite{jin2021temporal}.} \textcolor{black}{Note that the validation set is used for model and hyper-parameters selection.} All videos are sub-sampled to $5$ fps.
All frames of videos belong to one and only one phase label, and there are not any background frames.
{\bf (b) Cholec80.} The Cholec80 dataset~\cite{twinanda2016endonet} contains 80 videos of cholecystectomy surgeries, which are are recorded at $25$ fps and are annotated into $7$ surgical phases: \emph{Preparation, CalotTriangleDissection, ClippingCutting, GallbladderDissection, GallbladderRetraction, CleaningCoagulation, GallbladderPackaging}. The resolution of each frame is $1920 \times 1080$ or $854 \times 480$.
\textcolor{black}{We split the dataset into 40 training videos, $8$ validation videos and 40 testing videos, following the same setting in prior methods~\cite{jin2021temporal}.}
\textcolor{black}{Similar as M2CAI16, we also use the validation set for model and hyper-parameters selection.}
All videos of both M2CAI16 and Cholec80 datasets are sub-sampled into $5$ fps.
All frames in two datasets belong to one and only one phase label, and there are not any background frames.
\textcolor{black}{The datasets (M2CAI16 and Cholec80) we employed in this work are the largest public benchmark datasets, which have been widely used in many published surgical workflow recognition papers~\cite{jin2017sv,jin2020multi,jin2021temporal,yi2021not,czempiel2020tecno}. The datasets we used are very challenging and hard to be overfitting. The reasons are shown as follows: (a) The duration of each video is very long (around 30 minutes), and each of the dataset has around more than 170K frames, which is hard to over-fitting (note that our task is a frame-wise classification).  (b) The variance for videos and different phases is very large. For example, the Cholec80 datasets is collected by 13 surgeons which is very diverse. Furthermore, the standard deviation of duration for phase “Gallbladder dissection” is 551s. Each video may not have all phases and there is no obvious prior between two phases.}
%


\vspace{1.5mm}
\noindent{\bf Evaluation Metrics.}
We employ four commonly-used metrics,~\emph{i.e.}, accuracy
(AC), precision (PR), recall (RE), and Jaccard (JA) to evaluate the phase prediction accuracy. 
AC represents a video-level evaluation, which is defined as the percentage of correctly classified frames in the entire video.
Due to the imbalanced phases presented in videos, PR, RE, and JA refer to the phase-level evaluation, which is evaluated within each phase and then averaged over all the phases.
Specifically, we first compute PR, RE and JA of each phase by $\frac{|{GT} \cap {P}|}{|{P}|}$, $\frac{|{GT} \cap {P}|}{|{GT}|}$ and $\frac{|G T \cap P|}{|G T \cup P|}$, where $GT$ and $P$ refer to the ground-truth and prediction set, respectively.
Then, we \textcolor{black}{obtain the mean and the standard deviation of these scores over all phases and obtain the performance of the entire video.}
The evaluation protocols are the same with previous methods~\cite{jin2017sv,jin2020multi,jin2021temporal,czempiel2020tecno,yi2021not}.

\vspace{1.5mm}
\noindent{\bf Implementation Details.}
The model is built with Pytorch~\cite{pytorch} and is trained by 1 NVIDIA 3090 GPU. 
We use Adam~\cite{kingma2014adam} optimizer with the learning rate of $5 \times 10^{-4}$, decayed by 30 epochs, and \textcolor{black}{The model is totally trained for 100 epochs.}
\textcolor{black}{
We select the model with the highest performance on the validation, and reports its results on the test set.}
The number heads of segment-frame attention $N_{head}$ is set to $4$ and \textcolor{black}{the positional encoding is learned}, due to the variant duration of videos. %
In the frame-wise feature encoder, the number of the residual causal dilated temporal convolution layers (RCDL) is set to $11$.
In the segment-level hierarchical network, we set the number of RCDL to be $10$.
The dimension of the features,~\emph{i.e.}, $D$, is set to $64$.
We use the max-pooling as the temporal fusion layer, and we will evaluate the effect of different methods,~\emph{e.g.}, average-pooling or convolutions in Section~\ref{sec:ablation_param}.
We set the sizes of the kernel and the stride $k$ of the temporal fusion layers are both set to $7$, which achieves the best performance.
There is no overlapping when generating the segment-level features to avoid the same segment contains the frames from different phases.
The ablation study of $k$ is conducted in Section~\ref{sec:ablation_param}.
We set $\alpha$ and $\beta$ to be $1$ for both M2CAI16 and Cholec80, the detailed analysis is shown in Fig.~\ref{fig:beta&lamd}.
%

\subsection{Comparison with the State-of-the-Arts}

As shown in Table~\ref{table:m2cai} and Table~\ref{table:chole80}, we compare the proposed SAHC with the state-of-the-art approaches on the M2CAI16 and Cholec80 dataset, including PhaseNet~\cite{twinanda2016single}, SV-RCNet \cite{jin2017sv}, OHFM~\cite{yi2019hard}, TMRNet \cite{jin2021temporal}, Trans-SVNet \cite{gao2021trans}, TeCNO~\cite{czempiel2020tecno}.
\textcolor{black}{Note that ``ours" in Table~\ref{table:m2cai} and Table~\ref{table:chole80} indicate the model described in Fig.~\ref{fig:architecture},~\emph{i.e.}, using three scales with the segment-frame attention, and the implementation of the temporal fusion layer is max-pooling.}
Our method achieves \textcolor{black}{$4.8\%$} and \textcolor{black}{$3.0\%$} improvements over the prior state-of-the-arts on the M2CAI16 dataset and the Cholec80 dataset, respectively.
Notably, the improvements of our method are more significant in M2CAI16 than in Cholec80. This is because  M2CAI16 contains more ambiguous frames, as shown in Fig.~\ref{fig:motivation}(b), demonstrating the effectiveness of our method to address the main challenge in surgical video recognition. 

\textcolor{black}{
As described in Section~\ref{sec:datasets}, higher accuracy represents higher frame-level performance, while higher precision, recall, and Jaccard indicates higher phase-level accuracy. Since the imbalanced phases shown in videos, the phase-level performance is more reasonable for surgical video phase recognition.
Among three phase-level metrics, compared with precision ($\frac{|{GT} \cap {P}|}{|{P}|}$) and recall ($\frac{|{GT} \cap {P}|}{|{GT}|}$), Jaccard ($\frac{|G T \cap P|}{|G T \cup P|}$) measures how close the predictions of phase set to the ground-truth set are, which is more accurate and stricter.
From Table~\ref{table:m2cai} and Table~\ref{table:chole80}, we find that our method can achieve higher improvement of Jaccard than other metrics, which indicates that our model can produce less misclassification  prediction.
}
%
As shown in Fig.~\ref{fig:phase_vis}(c). ``GT" indicates ground-truth, ``Base" refers to the baseline model. The predictions of the baseline and our method have the similar Accuracy score. Compared to the baseline, the prediction of our method shows a higher Jaccard score, demonstrating that our model has a good capacity in refining erroneous predictions caused by ambiguous frames within a surgical phase.

\subsection{Ablation Study}
We ablate the proposed SAHC to evaluate the effectiveness of each component and analyze why they work.
Our baseline model is the network in Fig.~\ref{fig:architecture} without generating segment-level information and hierarchical segment-frame attention, denoted as ``Base'' for simplification. 
In other words, all output of the baseline model is at the same temporal scale,~\emph{i.e.}, frame-level. \textcolor{black}{Note that all models in this section are optimized and trained on the training dataset with corresponding losses independently.}
%
%
\begin{table}[t]
\centering
\caption{Ablation study on the segment-level information and the hierarchical consistency network on M2CAI16 dataset.}\smallskip
\label{table:msc}
\resizebox{1.0\columnwidth}{!}{
\begin{tabular}{c | c c c c }
   Methods  & Accuracy & Precision & Recall & Jaccard\\
\Xhline{1.5pt}
    Base &87.1 $\pm$ 9.0 &  87.7 $\pm$ 7.1  &  88.8 $\pm$ 6.3 &  76.6 $\pm$  8.9\\
    \hline
  $\{ \mathbf{F}^1 \}$ & \textcolor{black}{87.8 $\pm$ 7.3} &  \textcolor{black}{89.3 $\pm$ 6.5}  &  \textcolor{black}{89.2 $\pm$ 5.5} &  \textcolor{black}{82.5 $\pm$  6.8} \\
  $\{ \mathbf{F}^1, \mathbf{F}^2 \}$  & \textcolor{black}{88.6 $\pm$ 7.8} &  \textcolor{black}{90.9 $\pm$ 6.3}  &  \textcolor{black}{90.5 $\pm$ 5.2} &  \textcolor{black}{84.2 $\pm$  6.8}  \\
  $\{ \mathbf{F}^1, \mathbf{F}^2, \mathbf{F}^3 \}$  & \textcolor{black}{90.2 $\pm$ 8.8} &  \textcolor{black}{92.4 $\pm$ 6.2}  &  \textcolor{black}{91.3 $\pm$ 5.5} &  \textcolor{black}{84.8 $\pm$ 6.8} \\
\end{tabular}}
\end{table} 
\begin{table}[t]
\centering
\caption{Comparison of number of parameters,\textcolor{black}{computation cost and running time} of different models. \textcolor{black}{``Param'' indicates the number of parameters}. \textcolor{black}{``w/o SFA'' indicates our segment-level model without using SFA}.}\smallskip
\label{table:parameter}
\resizebox{1.0\columnwidth}{!}{
\begin{tabular}{c | c c c c}
   Methods  & Accuracy &  Param & \textcolor{black}{GFLOPS} & \textcolor{black}{Running time}\\
\Xhline{1.5pt}
    Base & 87.1 $\pm$ 9.0 &  26.8M & \textcolor{black}{1.60G}& \textcolor{black}{0.27s} \\
    \textcolor{black}{ Ours w/o SFA} &  \textcolor{black}{90.2 $\pm$ 8.8} &   \textcolor{black}{26.8M} & \textcolor{black}{1.83G}&  \textcolor{black}{0.29s} \\
    \textcolor{black}{Ours} &  \textcolor{black}{91.8 $\pm$ 8.1} &   \textcolor{black}{26.9M} &  \textcolor{black}{2.02G} & \textcolor{black}{0.31s} \\
\end{tabular}}
\end{table} 
\begin{table}[t]
\centering
\caption{Ablation study on the hierarchical segment-frame attention (SFA) on M2CAI16.}\smallskip
\label{table:abl_mssfa}
\resizebox{1.0\columnwidth}{!}{
\begin{tabular}{c | c c c c }
    Methods & Accuracy & Precision & Recall & Jaccard\\
\Xhline{1.5pt}
Segment w/o SFA & \textcolor{black}{90.2 $\pm$ 8.8} &  \textcolor{black}{92.4 $\pm$ 6.2}  &  \textcolor{black}{91.3 $\pm$ 5.5} &  \textcolor{black}{84.8 $\pm$ 6.8} \\ 
 \hline
 $\{ \mathbf{F}^1_{pos} \}$  &  \textcolor{black}{91.0 $\pm$ 7.9} & \textcolor{black}{92.8 $\pm$ 5.9}  &  \textcolor{black}{92.2 $\pm$ 4.8} &  \textcolor{black}{85.1 $\pm$  8.4}\\
 $\{ \mathbf{F}^1_{pos}, \mathbf{F}^2_{pos} \}$  & \textcolor{black}{91.6 $\pm$ 8.3} &  \textcolor{black}{93.0 $\pm$ 4.9}  &  \textcolor{black}{92.5 $\pm$ 5.5} &  \textcolor{black}{85.4 $\pm$  5.8} \\
 $\{ \mathbf{F}^1_{pos}, \mathbf{F}^2_{pos}, \mathbf{F}^3_{pos} \}$  &\textcolor{black}{{\bf 91.6 $\pm$ 7.8}} & \textcolor{black}{{\bf 93.5 $\pm$ 5.6}} & \textcolor{black}{{\bf 92.9 $\pm$  6.3}} & \textcolor{black}{{\bf 85.7 $\pm$  7.7}}\\
\end{tabular}}
\end{table} 
\begin{figure*}[t]
    \centering
    \includegraphics[width=1.8\columnwidth,height=0.3\textheight]{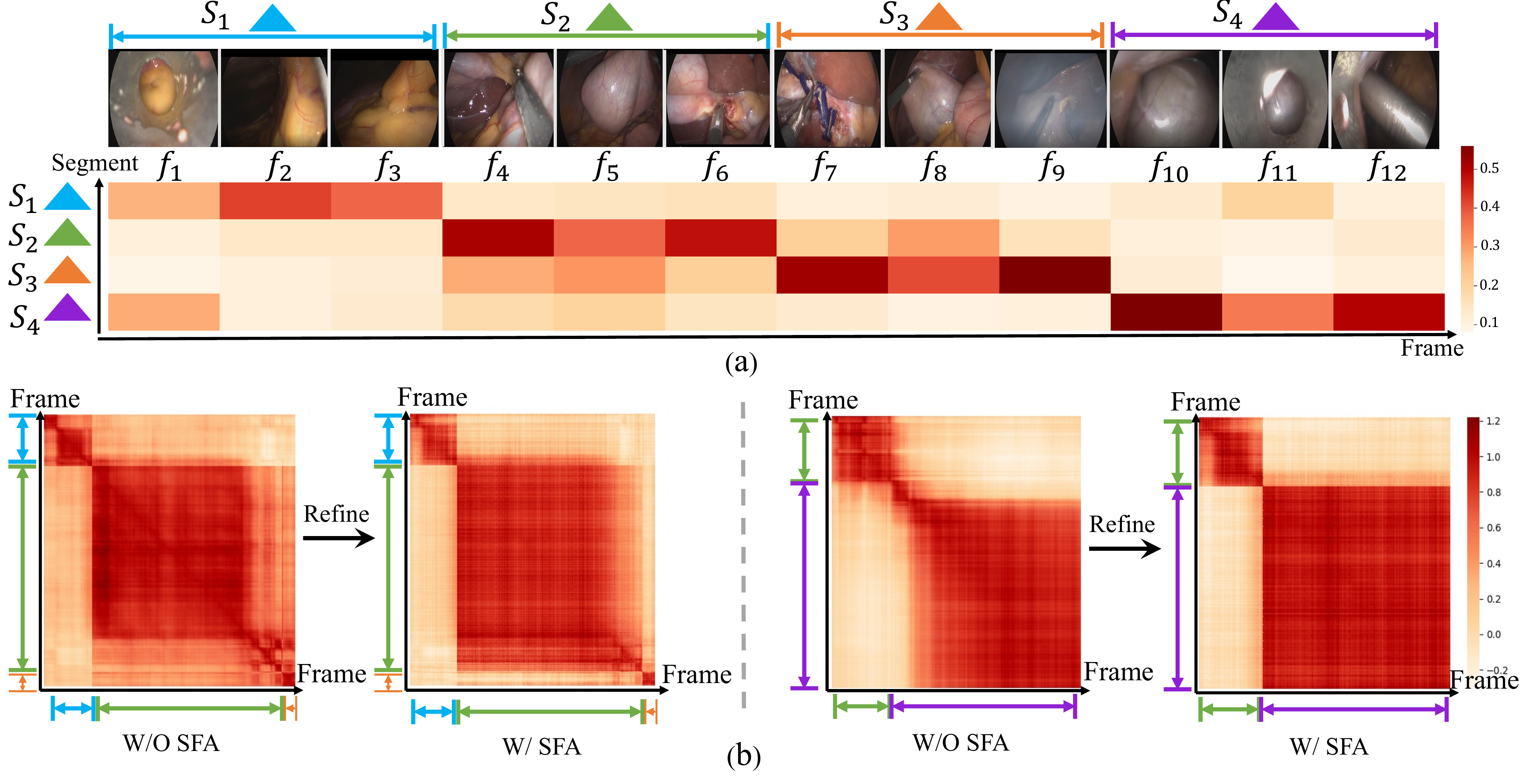}
    \caption{
    (a) Attention weight of the segment-frame attention (SFA) module. The horizontal axis represents the frame features, and the vertical axis represents the segment features.
    \textcolor{black}{In this figure, there are a total of 12 frames (\emph{i.e.}, $\{ f_i \}_{i=1}^{12}$) from four different categories of phases. Different colors indicate different phases,~\emph{e.g.}, $\{ f_i \}_{i=1}^{3}$ belong to the first phase (the blue one). The segment-level features (triangles with different colors) are generated by feeding three frames into SFE~(defined in Section~\ref{sec:sfe}),~\emph{i.e.}, $S_i = SFE(\{ f_{3*i-2}, f_{3*i-1}, f_{3*i}\})$.
    It is clear that the segment representations show the high similarity with the frame ones that sharing the same phase class.
    }
    %
    (b) The cosine similarity matrices of features of frames.
     The horizontal and vertical axis are both represent the frame features.
     The features in the same phase become more similar after the refinement of our method.
     The bi-direction arrows with different colors indicate the different phases.
    %
    %
    }
    \label{fig:att&sim}
\end{figure*}
\subsubsection{Effect of learning with segment-level semantics.}
\textcolor{black}{In this ablation study, we prove that high-level segment feature outperforms than the frame-level one. 
As shown in Fig.~\ref{fig:seglevel}, the performance of the model trained with segments,~\emph{i.e.}, ``3-scale" and ``5-scale", outperforms that of the model trained with frames,~\emph{i.e.},~``1-scale", on both four metrics.}

\subsubsection{Hierarchical Segment Information.}
\ \\
%
Table~\ref{table:msc} shows the importance of the segment-level information and the hierarchical consistency network on M2CAI16. $\{ \mathbf{a}, \mathbf{b}, \mathbf{c} \}$ indicates the consistency with $\mathbf{a}$, $\mathbf{b}$ and $\mathbf{c}$.
Compared ``Base'' and ``$\{ \mathbf{F}^1 \}$'', it is clear that the segment-level prediction can improve the performance from $87.1\%$ to \textcolor{black}{$87.8\%$} on accuracy.
We also find that the hierarchical consistency between segments and frames can achieve considerable improvements.
For example, with an additional scale of segment, \emph{e.g.}, $\mathbf{F}^2$, the model achieves \textcolor{black}{$1.5\%$} improvements in terms of accuracy.
With three-scale segments, the model reaches the best performance, \emph{i.e.}, \textcolor{black}{$90.2\%$} accuracy.
More scales of segments are not feasible in our dataset since the temporal size at $\mathbf{F}^3$ is too small to be downsampled.
More specifically, the average length of training videos in M2CAI16 is $2502$. After three times downsampling with $k=11$, the length of the final feature is only $1.8$, which is too small to cover whole phases in the video. Note that the number of phases in a video is generally from $3$ to $6$.
Furthermore, in order to prove that the improvement comes from the proposed hierarchical segment information, we also compare the number of parameters, \textcolor{black}{computation cost and running time} of \textcolor{black}{our} method with that of the baseline model, which is reported in Table~\ref{table:parameter}.
``Param'' indicates the number of parameters.
It is clear that our proposed method outperforms the baseline model with a clear margin,~\emph{i.e.}, \textcolor{black}{$4.7\%$}, while having the same number of parameters,~\emph{i.e.}, $26.8$M.
\textcolor{black}{Compared with the baseline, our model would bring extra 0.23G FLOPS.}
\textcolor{black}{Furthermore, the running time of training/inference of our model is $0.31$s for each video, only bringing extra 0.04s compared with the baseline model.} 
\begin{figure*}
    \centering
    \includegraphics[width=2\columnwidth,height=0.15\textheight]{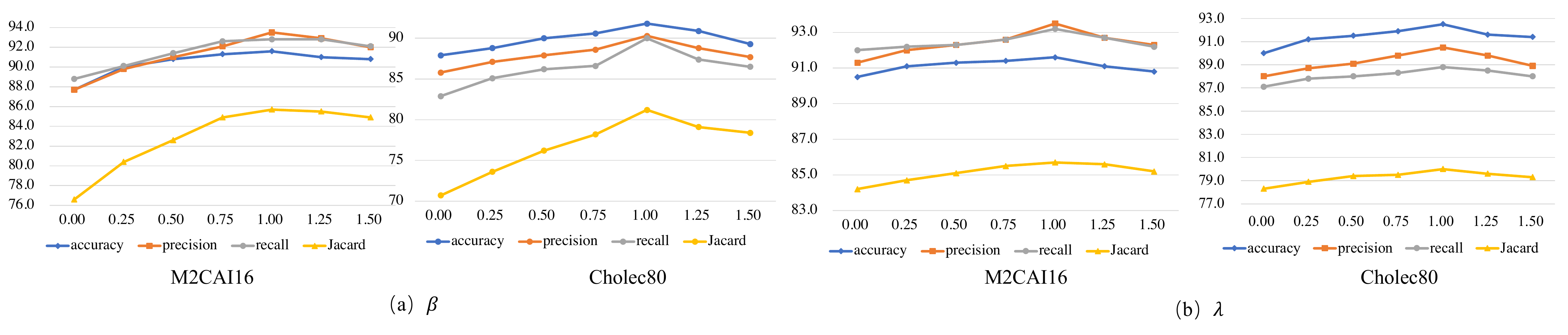}
    \caption{
  Analysis of (a) $\beta$ and (b) $\lambda$ on M2CAI16 and Cholec80.
  We show the results of Accuracy, Precision, Recall and Jaccard of models with different $\beta$ and $\lambda$.
    }
    \label{fig:beta&lamd}
\end{figure*}
\subsubsection{Segment-frame attention (SFA) module.}
\ \\
Table~\ref{table:abl_mssfa} shows the effectiveness of the hierarchical segment-frame attention (SFA) on M2CAI16 dataset.
$\{ \mathbf{F}^1_{pos}, \mathbf{F}^2_{pos}, \mathbf{F}^3_{pos} \}$ indicates SFA module between level 0 and 1, level 0 and 2, level 0 and 3, respectively.
From comparison of ``Segment w/o SFA ''and ``$\{ \mathbf{F}^1_{pos} \}$'', we can observe that using SFA between frames and one level of segment, the performance can achieve \textcolor{black}{$0.8\%$} improvement in terms of accuracy.
Moreover, using SFA between frames and hierarchical segments can achieve better performance, demonstrating the effectiveness of the proposed hierarchical network and SFA. 


To explore how segment-level information helps rectify erroneous predictions, we show the temporal attention weight of SFA in  Fig.~\ref{fig:att&sim}(a). 
Clearly, each high-level segment feature focuses on its neighboring frames,~\emph{i.e.}, the segment representations show the high similarity with the frame ones that sharing the same phase class. As a result, the features of frames (including ambiguous frames) belong to the same phase will be pulled together and refined to be consistent with the segment-level features. 
Fig.~\ref{fig:att&sim}(b) visualizes the cosine similarity of pair-wise frames sampled from the test dataset. Notably, by using SFA, the features of frames from the same phases would be similar, and vice versa. In other words, with SFA, our method can better recognize surgical phases.
For example, as shown in the right two heatmaps in Fig.~\ref{fig:att&sim}(b), the frames in the same phase,~\emph{e.g.}, the purple arrows, are become more similar when using SFA.

%
\begin{table}[t]
\centering
\caption{Comparison with the performance of SFE with different temporal fusion layers on M2CAI16. \textcolor{black}{``Conv", ``MP"  and ``AP" indicate convolution, max-pooling and average-pooling respectively.}}\smallskip
\label{table:abl_tfl}
\resizebox{1.0\columnwidth}{!}{
\begin{tabular}{c | c c c c }
   Methods  & Accuracy & Precision & Recall & Jaccard\\
\Xhline{1.5pt}
    Base &87.1 $\pm$ 9.0 &  87.7 $\pm$ 7.1  &  88.8 $\pm$ 6.3 &  76.6 $\pm$  8.9\\
    \hline
  Conv & \textcolor{black}{90.0 $\pm$ 7.4} &  \textcolor{black}{90.3 $\pm$ 6.5}  &  \textcolor{black}{91.0 $\pm$ 4.3} &  \textcolor{black}{84.4 $\pm$  7.5} \\
  MP  & \textcolor{black}{90.1 $\pm$ 7.5} &  \textcolor{black}{91.8 $\pm$ 5.5}  &  \textcolor{black}{91.1 $\pm$ 4.4} &  \textcolor{black}{84.7 $\pm$  7.8}  \\
  AP  & \textcolor{black}{90.5 $\pm$ 7.5} &  \textcolor{black}{92.4 $\pm$ 5.6}  &  \textcolor{black}{91.6 $\pm$ 4.5} &  \textcolor{black}{85.0 $\pm$  7.7} \\
\end{tabular}}
\end{table} 

\subsubsection{Ablation on Parameters}\label{sec:ablation_param}
\begin{figure*}[t]
    \centering
    \includegraphics[width=2.0\columnwidth,height=0.22\textheight]{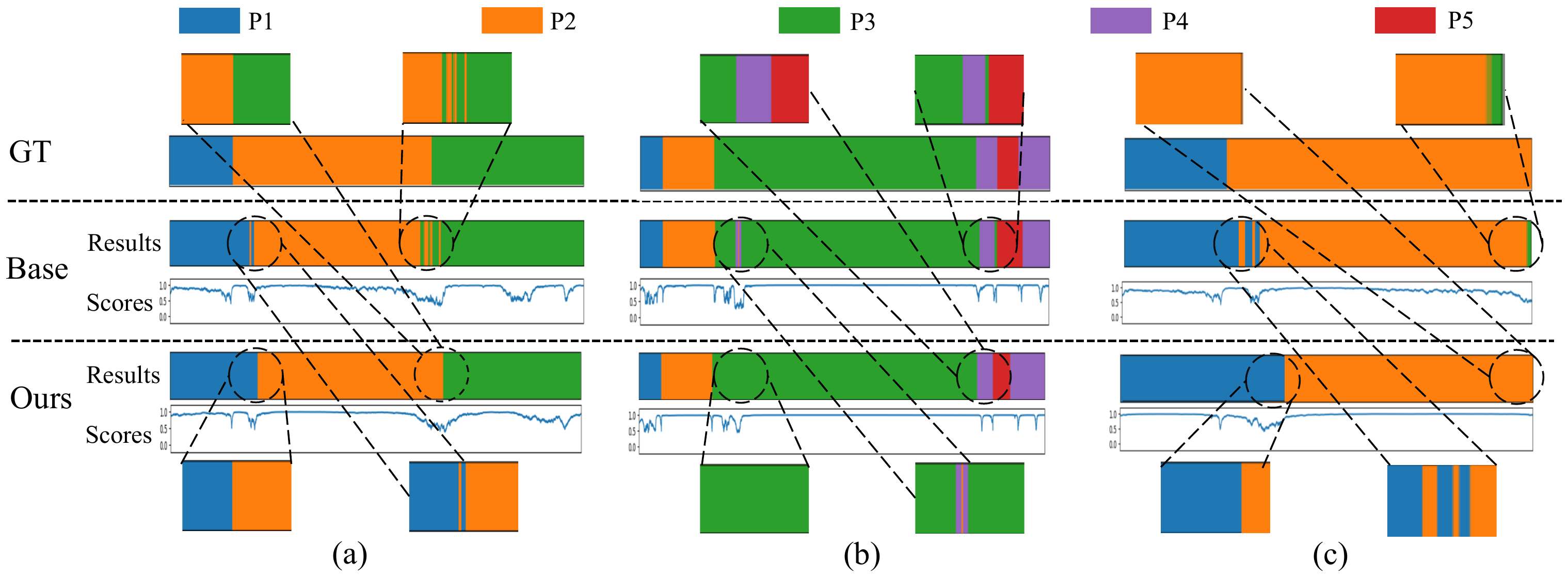}
    \caption{
    Prediction results of three video examples from M2CAI16 dataset.
    ``P1-P5" indicates the phase 1 to phase 5.
    ``GT'' refers to the ground-truth and ``Base'' refers to baseline. 
    The top row and the bottom row in Base and ours are the prediction results and the prediction scores, respectively.
    We scale the temporal axes of all videos for better visualization.
    }
    \label{fig:phase_vis}
\end{figure*}
\noindent{\bf Comparison with different temporal fusion layers.}
As shown in Table~\ref{table:abl_tfl}, we ablate the performance of different temporal fusion layers, \emph{i.e.}, convolution (Conv), max-pooling (MP) and average-pooling (AP).
Note that in this ablation study, we just aim to compare the effect of different temporal fusion layers, and do not use the segment-frame attention (SFA) module.
We find that AP is better than MP and Conv,~\emph{e.g.}, achieving \textcolor{black}{$90.5\%$} accuracy, \textcolor{black}{$0.4\%$} over than MP. This is because AP would encourage the overall frames in the kernel windows to be consistent with the high-level segment one. Hence, the features of the ambiguous frame would be pulled together with its corresponding high-level segments.

\vspace{1.5mm}
\noindent{\bf Comparison of SFA with different models.}
%
In Table~\ref{table:abl_sfa}, we compare the performance of SFA with Base and different temporal fusion layers,~\emph{i.e.}, Conv MP and AP.
``Conv", ``MP" and ``AP "indicate the convolution, max-pooling and average-pooling respectively.
SFA is not very impressive without the segment-level information (Base),~\emph{i.e.}, only $0.6\%$ improvement on Base.
However, SFA can bring considerable improvement on the model with segment-level information.
For example, the temporal fusion layer with Conv achieves \textcolor{black}{$91.2\%$} accuracy, outperforms \textcolor{black}{$3.5\%$} over the baseline model.
The result demonstrates that SFA can capture the relations between high-level segments and low-level frames, which can help refine erroneous predictions in low-level frames.
Furthermore, we also find that use AP as the temporal fusion layer bring the largest improvement for SFA, compared with Conv and MP.
For instance, we obtain \textcolor{black}{$91.6\%$} accuracy on the temporal fusion layer with AP, \textcolor{black}{$0.4\%$} and \textcolor{black}{$0.3\%$} improvement over that with Conv and MP respectively.

\begin{table}[t]
\centering
\caption{Ablation study on the performance of SFA with different models on M2CAI16.}\smallskip
\label{table:abl_sfa}
\resizebox{1.0\columnwidth}{!}{
\begin{tabular}{c | c c c c }
   Methods  & Accuracy & Precision & Recall & Jaccard\\
\Xhline{1.5pt}
    Base &87.1 $\pm$ 9.0 &  87.7 $\pm$ 7.1  &  88.8 $\pm$ 6.3 &  76.6 $\pm$  8.9\\
    \hline
   SFA w/ Base & \textcolor{black}{87.5 $\pm$ 7.0} &  \textcolor{black}{88.2 $\pm$ 5.4}  &  \textcolor{black}{89.0 $\pm$ 6.0} &  \textcolor{black}{77.0 $\pm$  7.0}\\
  \hline
  SFA w/ Conv & \textcolor{black}{91.2 $\pm$ 7.3} &  \textcolor{black}{92.5 $\pm$ 5.5}  &  \textcolor{black}{92.7 $\pm$ 4.5} &  \textcolor{black}{85.3 $\pm$  7.2} \\
 SFA w/ MP & \textcolor{black}{91.3 $\pm$ 7.5} &  \textcolor{black}{93.2 $\pm$ 5.5}  &  \textcolor{black}{92.9 $\pm$ 4.5} &  \textcolor{black}{85.6 $\pm$  7.7}  \\
   SFA w/ AP  & \textcolor{black}{{\bf 91.6 $\pm$ 7.8}} & \textcolor{black}{{\bf 93.5 $\pm$ 5.6}} & \textcolor{black}{{\bf 92.9 $\pm$  6.3}} & \textcolor{black}{{\bf 85.7 $\pm$  7.7}}\\
\end{tabular}}
\end{table}

\begin{table}[!t]
\centering
\caption{Comparison with different kernel sizes $k$ of the temporal fusion layer on M2CAI16.}\smallskip
\label{table:abl_resize}
\resizebox{0.9\columnwidth}{!}{
\begin{tabular}{c | c c c c }
   Size  & Accuracy & Precision & Recall & Jaccard\\
\Xhline{1.5pt}
    $1/1$ & 87.1 $\pm$ 9.0 &  87.7 $\pm$ 7.1  &  88.8 $\pm$ 6.3 &  76.6 $\pm$  8.9\\
    $1/3$ & \textcolor{black}{91.2 $\pm$ 7.4} & \textcolor{black}{93.0 $\pm$ 5.9}  &  \textcolor{black}{92.5 $\pm$ 4.0} &  \textcolor{black}{84.5 $\pm$  7.8}\\
    $1/5$ & \textcolor{black}{91.5 $\pm$ 7.7} &  \textcolor{black}{93.2 $\pm$ 6.5}  &  \textcolor{black}{92.5 $\pm$ 4.4} &  \textcolor{black}{85.4 $\pm$  7.4} \\
    $1/7$  & \textcolor{black}{{\bf 91.6 $\pm$ 7.8}} & \textcolor{black}{{\bf 93.5 $\pm$ 5.6}} & \textcolor{black}{{\bf 92.9 $\pm$  6.3}} & \textcolor{black}{{\bf 85.7 $\pm$  7.7}}\\
    $1/9$  & \textcolor{black}{91.2 $\pm$  7.7} &  \textcolor{black}{93.2 $\pm$ 5.6}  &  \textcolor{black}{92.6 $\pm$ 6.5} &  \textcolor{black}{85.3 $\pm$  7.7} \\
  $1/11$ & \textcolor{black}{91.1 $\pm$ 7.7} &  \textcolor{black}{93.3 $\pm$ 5.5}  &  \textcolor{black}{92.3 $\pm$ 6.4} &  \textcolor{black}{85.2 $\pm$  7.6}  \\
\end{tabular}}
\end{table} 

\vspace{1.5mm}
\noindent{\bf Comparison with different kernel sizes $k$.}
In Table~\ref{table:abl_resize}, we analyze different kernel sizes $k$ (see Eq.~\ref{e:tfl}) in the temporal fusion layer.
$1/\{1,3,5,7,9,11 \}$ indicates that we set $k$ to be $\{1,3,5,7,9,11 \}$ respectively.
Note that $k=1$ indicates that the baseline model without the segment-level information.
It is clear that with $k>1$,~\emph{i.e.}, the segment-level information, the model achieve significant improvement.
We can observe that too small and too large kernel size would hurt the performance.
This is due to that too small kernel size may extract very local segment-level information, which can not capture high-level information to refine the ambiguous frames,~\emph{i.e.}, setting $k$ to $3$ only achieves \textcolor{black}{$91.2\%$} accuracy.
On the other hand, too large kernel size may include the frames from different phases into one high-level segment, leading to mistakes in generating segments,~\emph{i.e.}, setting $k$ to $11$ only achieves \textcolor{black}{$91.1\%$} accuracy.
%
In our experiments, we find that $1/7$ is the best reduction rate, which achieves \textcolor{black}{$91.6\%$} accuracy, \textcolor{black}{$0.4\%$} and \textcolor{black}{$0.1\%$} over that of $1/9$ and $1/5$ respectively. 

\vspace{1.5mm}
\noindent{\bf Analysis on hyper-parameters $\beta$ and $\lambda$.}
Fig.~\ref{fig:beta&lamd} shows the model performance with different values of $\beta$ in Eq.~\ref{E:seg} and $\lambda$ in Eq.~\ref{E:all}.
$\beta$ controls the weight of the frame-wise prediction and the segment-wise prediction.
Setting $\beta$ to zero indicates the model without the consistency between predictions of frames and segments, which shows the limited performance,~\emph{i.e.}, only obtaining the accuracy score of $87.7\%$ on M2CAI16.
%
In the experiments, we find that setting $\beta$ to $1.0$, our method achieves the best performance,~\emph{i.e.}, achieving \textcolor{black}{$91.6\%$} accuracy on M2CAI16, \textcolor{black}{$3.9\%$} over that setting $\beta$ to $0.0$.
$\lambda$ controls the importance of smoothing, which regularizes the model to predict smoothing results.
It is clear that combining with the smoothing regularization, the model can achieve the better performance.
For example, the accuracy of the model setting $\lambda$ to $0.50$ outperforms \textcolor{black}{$0.8\%$} over that of the model without $\mathcal{L}_{smooth}$,~\emph{i.e.}, $\lambda=0.00$.
We find that setting $\lambda$ to $1.0$ achieve the best performance on both M2CAI16 and Cholec80 datasets,~\emph{i.e.},\textcolor{black}{$91.6\%$} and \textcolor{black}{$91.8\%$} respectively.


\subsection{Qualitative Analysis and Discussion}
Fig.~\ref{fig:phase_vis} shows the qualitative results of our method.
We scale the temporal axes of three videos,~\emph{i.e.}, (a)-(c), for better visualization.
As shown in Fig.~\ref{fig:phase_vis} (a)-(b), our method predicts higher frame-wise accuracy, especially in the boundaries between two different phases.
Furthermore, we can notice that ``Base'' makes many mistakes,~\emph{i.e.}, classifying ambiguous frames within different phases.
For example, some frames within P3 are misclassified into P4 and some ones within P4 are misclassified into P3, as shown in Fig.~\ref{fig:phase_vis}(b).
On the contrary, our proposed method can predict more robust and smooth results.
Note that our methods predict higher Jaccard scores as shown in Fig.~\ref{fig:phase_vis}(c).
Although the predictions of the baseline model and ours show the similar frame-wise accuracy, our methods present much higher phase-wise Jaccard scores,~\emph{i.e.}, more smooth results.
From the comparison of Fig.~\ref{fig:phase_vis}(b) and (c), we can find that our method can solve the ambiguous prediction inside phases very well.

However, both the baseline model and our method generate the misalignment boundaries when transits on phase to another one, as shown in Fig.~\ref{fig:phase_vis}(c).
We believe that this may be due to the noise annotations, since it is difficult to determine which frame is the precise boundary, and this process is subjective.
In the future, we may use some uncertainty analysis methods~\cite{peterson2019human,maddox2019simple,li2021uncertainty,wang2022less} to alleviate this problem.
\textcolor{black}{Furthermore, in real life, surgeons may sometime have to redo a phase which leads to more complicated surgical activities than the current public datasets. In the future, we will collect the surgical videos by ourselves to develop algorithms for this situation. Moreover, we will develop a more flexible frame-level module and design an attention module to capture the relationships among different or recurrent phases.}



\section{Conclusion}

This paper presents a novel segment-attentive hierarchical consistency network (SAHC) for surgical phase recognition from videos. 
Unlike previous methods, our key idea is to explore the segment-level semantics and use it to refine the erroneous predictions caused by ambiguous low-level frames.
SAHC consists of two innovative modules: a segment-level hierarchical consistency network to generate high-level semantic-consistent segments and a segment-frame attention (SFA) module to better reflect high-level segment information to low-level frames. 
Our method achieves \textcolor{black}{improved estimates of performance} on two public surgical video recognition datasets. 
Ablation study demonstrates the effectiveness of the proposed segment-level hierarchical consistency network and SFA module.

%


\bibliography{ref.bib}

\bibliographystyle{IEEEtran.bst}

\end{document}